\documentclass[conference]{IEEEtran}

\usepackage{times}

\let\labelindent\relax
\usepackage{enumitem}

\usepackage{xcolor} 
\usepackage{svg} 
\usepackage{graphicx} 
\usepackage{tabularx} 

\usepackage[numbers]{natbib}
\usepackage{multicol}

\usepackage{hyperref}
\hypersetup{bookmarks=true}

\usepackage{amsmath}

\usepackage{amsfonts}

\usepackage{stfloats}
\usepackage{float}

\newcommand\scalemath[2]{\scalebox{#1}{\mbox{\ensuremath{\displaystyle #2}}}}

\usepackage{listings}
\definecolor{codegreen}{rgb}{0,0.6,0}
\definecolor{codegray}{rgb}{0.5,0.5,0.5}
\definecolor{codepurple}{rgb}{0.58,0,0.82}
\definecolor{backcolour}{rgb}{0.98,0.98,0.95}

\usepackage{subcaption}
\usepackage[font={small}]{caption}
\usepackage[font={small}]{subcaption}

\lstdefinestyle{blockstyle}{
    backgroundcolor=\color{backcolour},   
    commentstyle=\color{codegreen},
    keywordstyle=\color{magenta},
    numberstyle=\tiny\color{codegray},
    stringstyle=\color{codepurple},
    basicstyle=\ttfamily\footnotesize,
    breakatwhitespace=false,         
    breaklines=true,                 
    captionpos=none,                    
    keepspaces=true,                 
    numbers=none,                    
    numbersep=5pt,                  
    showspaces=false,                
    showstringspaces=false,
    showtabs=false,                  
    tabsize=2
}
\begin{document}

\title{SymForce: Symbolic Computation and \\Code Generation for Robotics}


\author{

Hayk Martiros, Aaron Miller, Nathan Bucki, Bradley Solliday, Ryan Kennedy,  \\
Jack Zhu, Tung Dang, Dominic Pattison, Harrison Zheng, Teo Tomic, Peter Henry,   \\
Gareth Cross, Josiah VanderMey, Alvin Sun, Samuel Wang, Kristen Holtz \\

\textbf{Skydio, Inc.}
}

\maketitle


\begin{abstract}

We present SymForce, a library for fast symbolic computation, code generation, and nonlinear optimization for robotics applications like computer vision, motion planning, and controls. SymForce combines the development speed and flexibility of symbolic math with the performance of autogenerated, highly optimized code in C++ or any target runtime language. SymForce provides geometry and camera types, Lie group operations, and branchless singularity handling for creating and analyzing complex symbolic expressions in Python, built on top of SymPy. Generated functions can be integrated as factors into our tangent-space nonlinear optimizer, which is highly optimized for real-time production use. We introduce novel methods to automatically compute tangent-space Jacobians, eliminating the need for bug-prone handwritten derivatives. This workflow enables faster runtime code, faster development time, and fewer lines of handwritten code versus the state-of-the-art. Our experiments demonstrate that our approach can yield order of magnitude speedups on computational tasks core to robotics. Code is available at \href{https://github.com/symforce-org/symforce}{https://github.com/symforce-org/symforce}.

\end{abstract}

\IEEEpeerreviewmaketitle

\section{Introduction}

\setlength{\belowcaptionskip}{-15pt}

SymForce is a symbolic computation and code generation library that combines the development speed and flexibility of symbolic mathematics in Python with the performance of autogenerated, highly optimized code in C++ or any target runtime language. SymForce makes it possible to code a problem once in Python, experiment with it symbolically, generate optimized code, and then run highly efficient optimization problems based on the original problem definition.

Our approach was motivated by developing algorithms for autonomous robots at scale at Skydio, where performance and code maintainability are crucial for use cases like computer vision, state estimation, motion planning, and controls.

SymForce builds on top of the symbolic manipulation capabilities of the SymPy library \cite{10.7717/peerj-cs.103}. Symbolic math allows for rapid understanding, interactive analysis, and symbolic manipulations like substitution, solving, and differentiation. SymForce adds symbolic geometry and camera types with Lie group operations, which are used to autogenerate fast runtime classes with identical interfaces. By using one symbolic implementation of any function to generate runtime code for multiple languages, we improve the iteration cycle, minimize the chance of bugs, and achieve performance that matches or exceeds state-of-the-art approaches with no specialization.

\begin{figure}[H]
  \includegraphics[width=\columnwidth]{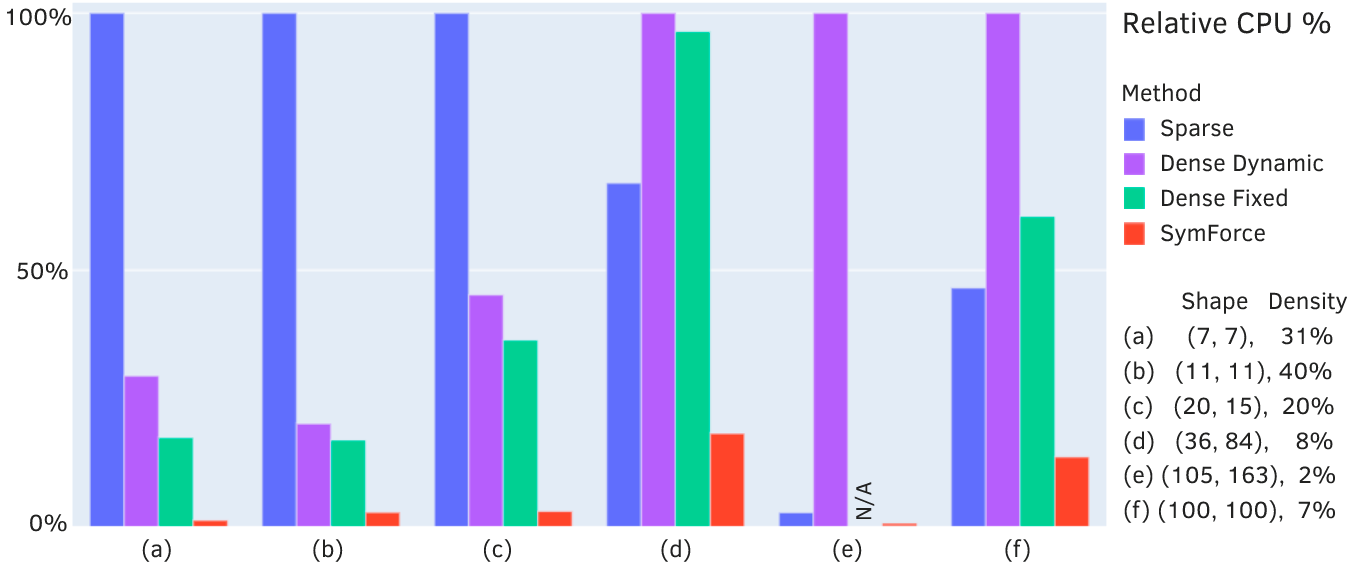}
  \caption{SymForce outperforms sparse and dense matrix multiplication with the Eigen library \cite{eigenweb} on our task in \ref{sec:matmul-experiment}, by sharing common subexpressions and leveraging sparsity with no runtime overhead.}
  \label{fig:bars}
  \vspace{0.7em}
\end{figure}

A significant advantage to our approach is not having to implement, test, or debug any Jacobians. In the robotics domain, correct and efficient computation of derivatives is critical. The prevalent approach is to hand-write Jacobians in C++ or CUDA for a core set of operations and rely on automatic differentiation to chain them together \cite{autodiff-review}. SymForce introduces novel methods to automatically compute tangent-space Jacobians of functions that operate on Lie groups, avoiding bug-prone handwritten derivatives. As a result, our approach avoids dynamic memory allocation and chain ruling at runtime. SymForce often dramatically outperforms standard approaches by flattening code across expression graphs, sharing subexpressions, and taking advantage of sparsity. We also introduce a novel method for preventing singularities without introducing branches, which has a key benefit to performance.

In summary, our key contributions are:

\begin{itemize}

    \item A free and open-source library with:
    
    \begin{itemize}

        \item Symbolic implementations of geometry and camera types with Lie group operations, and fast runtime classes with identical interfaces,
    
        \item Code generation for turning arbitrary symbolic functions into structured and fast runtime functions,
    
        \item A fast tangent-space optimizer in C++ and Python,
    
        \item Highly performant, modular, and extensible code,
    
    \end{itemize}

    \item Novel contributions for automatically computing tangent-space Jacobians, avoiding all handwritten derivatives,

    \item A novel method for avoiding singularities in complex expressions without introducing branching,

    \item An exposition of the speed benefits afforded by flattening computation across functions and matrix multiplications, especially for outperforming automatic differentiation.

\end{itemize}

\vspace{0.5em}
\smallskip
\smallskip

\begin{figure*}[t]
  \begin{center}
  \includegraphics[width=\textwidth]{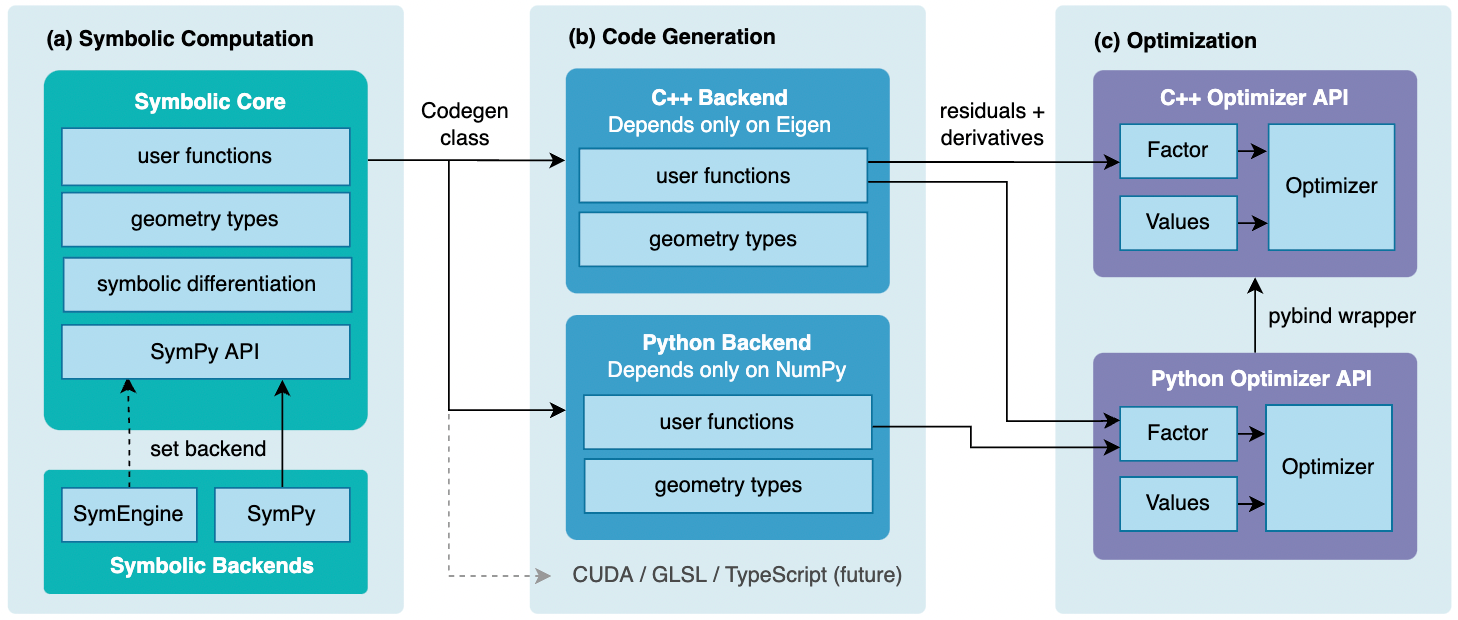}
  \end{center}
  \caption{\textbf{SymForce architecture diagram} - Symbolic expressions are written in Python, where they are easy to understand and debug with symbolic manipulation tools. Fast runtime code is autogenerated from these expressions, which can then be used as standalone functions or factors in our nonlinear optimizer. Together these three components provide a smooth workflow from prototypes to production code.}
  \label{fig:architecture}
\end{figure*}

The paper is organized as follows: We review related work in Sec. \ref{sec:related-work}. In Sec. \ref{sec:architecture} we present the major components of SymForce as shown in Figure~\ref{fig:architecture}: symbolic computation with geometry and camera types, code generation, and optimization.  Sec. \ref{sec:speed-advantages} highlights why SymForce is often faster than alternatives. Sec. \ref{sec:symbolic-differentiation} discusses symbolic differentiation on Lie groups, and Sec. \ref{sec:singularity-handling} describes our approach to branchless singularity handling. Limitations are discussed in Sec. \ref{sec:limitations}, experiments in Sec. \ref{sec:experiments}, and we conclude in Sec. \ref{sec:conclusion}.

\section{Related Work}
\label{sec:related-work}

\textit{\textbf{Symbolic Math}} -- Libraries that manipulate symbolic expressions, such as Maple \cite{maple}, Mathematica \cite{Mathematica}, and MATLAB’s symbolic toolbox \cite{matlabsymbolic}, have existed for decades. Symbolic math libraries that generate fast code are used in several niches of robotics \cite{10.1145/1644001.1644007} \cite{10.1115/DETC2013-13470} \cite{frost}, but are not general or widespread. SymPy \cite{10.7717/peerj-cs.103} is noteworthy for being open-source, lightweight, and written in Python, which allows users to modify it and leverage a large ecosystem of libraries. However, it can scale poorly to complex expressions. SymEngine \cite{symengine} is a C++ backend that supports the SymPy API while being up to two orders of magnitude faster \cite{symengine-speedups}. These libraries also provide routines for code generation and common subexpression elimination (CSE). SymForce builds on the capabilities of SymPy and SymEngine by adding essential geometry and vision types, Lie group operations, tools for handling singularities, structured code generation, and integration with optimization tools. These additions make SymForce a significantly more complete tool for complex robotics tasks.

\textit{\textbf{Nonlinear optimization}} -– Optimization libraries like GTSAM \cite{dellaert2012factor}, Ceres \cite{ceres-solver}, and g2o \cite{5979949} provide useful tools for efficiently formulating and minimizing cost functions, especially for nonlinear least-squares problems found in the field of robotics and computer vision. SymForce follows the factor graph formulation used in GTSAM, but residual functions are autogenerated from symbolic expressions. Relative to alternatives, our optimizer has faster performance and lower memory overhead for many tasks. In addition, our approach eliminates the need for handwritten derivatives and avoids the runtime overhead of applying the chain rule to Jacobians during automatic differentiaton.

\textit{\textbf{Geometric Lie Groups}} -- Rotations, poses, and camera models are central to robotics code, and getting efficient and correct tangent-space Jacobians is a common task for many optimization use cases. GTSAM defines geometry types and their Lie calculus in C++ and wraps them in Python. Sophus \cite{sophus} uses SymPy implementations to generate some derivative expressions. Manif \cite{Deray-20-JOSS} is a C++ library with handwritten Lie group operations. The wave\_geometry library \cite{koppel2018manifold} provides expression template-based automatic differentiation of Lie groups in C++. LieTorch \cite{DBLP:journals/corr/abs-2103-12032} implements Lie groups in PyTorch. SymForce is inspired by GTSAM and Sophus. The definitions of our geometry types are symbolic, with improved naming and consistency over Sophus and no hardcoded derivatives. SymForce autogenerates runtime classes that resemble the GTSAM variants in C++, but have faster performance and do not require maintaining handwritten code.





\textit{\textbf{Automatic Differentiation}} -- Libraries like PyTorch \cite{NEURIPS2019_9015}, TensorFlow \cite{tensorflow2015-whitepaper}, and JAX \cite{jax2018github} can build up computation graphs by tracing python code and applying automatic differentiation (AD) to compute gradients. These libraries often have high overhead for the small input dimensions that SymForce is built to tackle (tens to hundreds of variables). They also perform poorly for second-order optimization techniques common in robotics, because they are primarily targeted at computing gradients and not Hessians, and have poor performance for sparse matrices. In contrast, SymForce can compute Jacobians and Hessians of complex expressions with no operational overhead, making it suitable for use on resource-constrained embedded platforms. In addition, SymForce avoids the need for matrix multiplication at runtime that is needed for AD. Julia provides several AD libraries \cite{juliadiff}, but does not reach the performance of C++ and lacks the broader ecosystem of Python. Numba \cite{numba} can accelerate arbitrary Python functions, but does not compute derivatives.

\section{Architecture}
\label{sec:architecture}

This section describes the major components and workflows of the SymForce library, as summarized in Figure~\ref{fig:architecture}. Outside of this paper, the online documentation provides many tutorials and examples to gain a practical understanding of the library.

\subsection{Symbolic Computation}
\label{subsec:symbolic-computation}

SymForce provides tools for building and analyzing complex symbolic expressions in Python, by extending the SymPy API. There is a complete separation between code structure and performance, allowing the user to encapsulate their code without sacrificing performance.

\begin{figure*}[t!]
  \begin{center}
  \includegraphics[width=\textwidth]{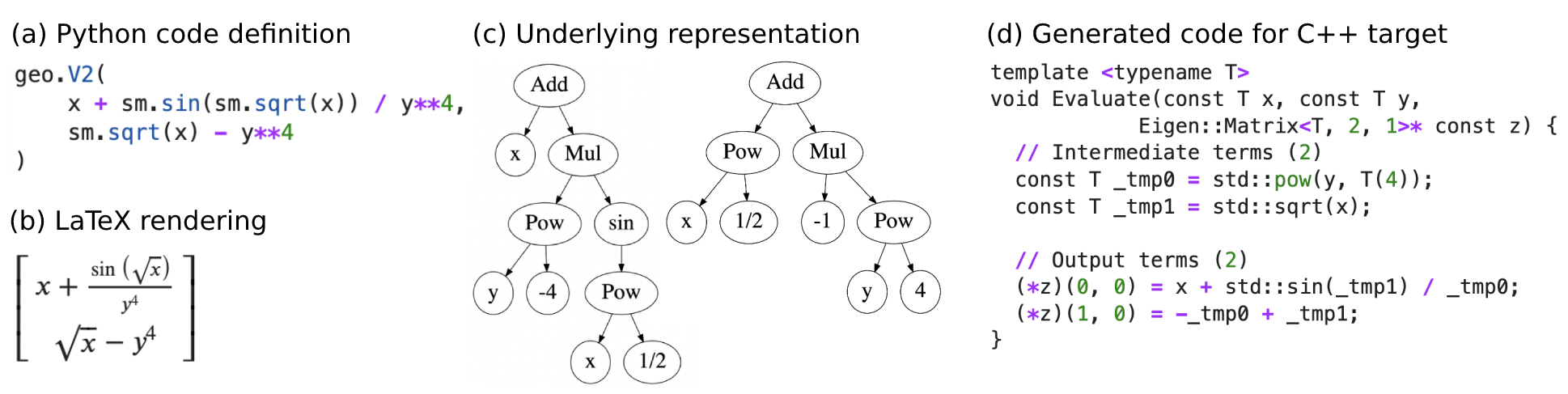}
  \end{center}
  \caption{\textbf{Symbolic expressions} - Representation of a simple two-vector symbolic expression (A) as defined in Python code, (B) automatically displayed as \LaTeX, (C) the underlying representation as a sequence of expression trees, (D) generated C++ and Eigen code. Complex expressions can contain hundreds of thousands of operations.}
  \label{fig:expression}
\end{figure*}

Symbolic computation centers around manipulation of mathematical expressions as algebraic instead of numerical quantities. Functions, symbols, and literals are stored as classes in code that can be traversed and transformed with symbolic substitution, simplification, differentiation, and solving. These tools allow users to interactively study and refine their functions. Fig. \ref{fig:expression} shows multiple forms of a simple expression.

SymForce supports two symbolic backends -- SymPy and SymEngine. SymEngine is a C++ implementation compatible with the SymPy API, but is dramatically faster for manipulating large expressions. It is the default choice.

\subsubsection{Geometry and Camera Types}
\label{subsubsec:geo-types}

SymForce implements several core types used in robotics, such as  matrices, rotations, poses, camera models, noise models, and barrier functions. These types have symbolic implementations defined as Python classes. Fast runtime classes with identical interfaces are autogenerated for target languages.

The generated classes do not depend on the rest of SymForce and can be deployed as standalone tools with minimal dependencies. For example, the generated C++ classes depend only on Eigen and the Python types only on NumPy. In C++ they are templated on the scalar type, and require zero dynamic memory allocation, making them suitable for embedded environments. The autogenerated runtime classes are functional, with readable interfaces and excellent test coverage.

\subsubsection{Lie Group Operations}
\label{subsubsec:lie-group-operations}

To allow generic code to operate on all geometry types we use a concepts (or traits) mechanism \cite{10.1145/1167515.1167499} inspired by GTSAM. Using concepts instead of inheritance allows generic programming with external types such as language-defined scalar and sequence types, or NumPy and Eigen matrices. It also makes it easy to extend or add new types, both within SymForce and in user code.

\noindent There are three primary concepts defined in SymForce:

\begin{enumerate}
\item \textit{\textbf{StorageOps}} allow all types to be converted to and from a sequence of scalars which comprise the internal representation of the type. For example, the SO(3) type is represented by a quaternion, whose storage contains 4 scalars. It also supports common tasks like substitution, numerical evaluation, and simplification.
\item \textit{\textbf{GroupOps}} register types as mathematical groups. A group type must have an associative composition operation, an identity element, and an inverse.
\item \textit{\textbf{LieGroupOps}} register types as Lie groups, meaning that in addition to being groups, they are differentiable manifolds with operations to go from group elements to perturbations in a local Euclidean space (tangent space).
\end{enumerate}

In Python, this mechanism is implemented with dynamic dispatch, for example \lstinline{ops.GroupOps.inverse(element)}.  In C++, this is done via template specialization, for example \lstinline{sym::GroupOps<sym::Pose3f>::Inverse()}.

The correctness of group and Lie group operations, and their Jacobians, have been rigorously tested both symbolically and numerically. See Sec. \ref{sec:symbolic-differentiation} and \ref{sec:singularity-handling} for details on our approach.

\subsection{Code Generation}
\label{subsec:code-generation}

Generation of fast runtime code from symbolic expressions is the core of SymForce. Our \lstinline{Codegen} class is the primary tool for all code generation tasks. It uses SymPy's code printers and adds support for struct types, Eigen / NumPy matrices, and our geometry and camera types. The resulting function contains flattened native code, but with a structured and human-readable interface that integrates nicely with our geometry types and optimization framework.

C++ is the most important code generation backend for SymForce. Generated C++ functions are templated on the scalar type with support for \lstinline{float}, \lstinline{double}, and \lstinline{std::complex}. However, we make it simple to add code generation backends targeting new languages, leveraging our template system based on the \lstinline{jinja} Python library \cite{jinja}.



A critical step in code generation is common subexpression elimination (CSE), which is the process of traversing a symbolic expression to find duplicate intermediate terms and pulling them out into temporary variables that are computed only once. CSE results in enormous efficiency gains, as described in Sec. \ref{subsec:function-flattening} and \ref{subsec:sparsity-exploitation}.

Take the following symbolic function that computes the residual of the position of a point expressed in a local frame to a point expressed in the world frame:

\begin{lstlisting}[language=Python]
def point_residual(
    world_T_local: geo.Pose3,
    world_point: geo.Vector3,
    local_point: geo.Vector3
) -> geo.Vector3:
    return world_point - world_T_local * local_point
\end{lstlisting}

Passing this symbolic function into our \lstinline{Codegen.function} method will generate a native function. An example for our C++ backend is shown in Fig.~\ref{fig:point-residual-code}.

\begin{figure}
\begin{lstlisting}[language=C++]
template <typename Scalar>
Eigen::Matrix<Scalar, 3, 1> PointResidual(
    const sym::Pose3<Scalar>& world_T_local,
    const Eigen::Matrix<Scalar, 3, 1>& world_point,
    const Eigen::Matrix<Scalar, 3, 1>& local_point
) {
  // Total ops: 53

  const Eigen::Matrix<Scalar, 7, 1>& _world_T_local = world_T_local.Data();

  // Common subexpressions (11)
  const Scalar _tmp0 = -2 * std::pow(_world_T_local[1], Scalar(2));
  // _tmp1 through _tmp9 omitted for brevity
  const Scalar _tmp10 = _tmp4 * _world_T_local[3];

  Eigen::Matrix<Scalar, 3, 1> _res;
  _res(0, 0) = -_world_T_local[4] - local_point(0, 0) * (_tmp0 + _tmp1) -
               local_point(1, 0) * (-_tmp3 + _tmp5) - local_point(2, 0) * (_tmp6 + _tmp7) +
               world_point(0, 0);
  _res(1, 0) = -_world_T_local[5] - local_point(0, 0) * (_tmp3 + _tmp5) -
               local_point(1, 0) * (_tmp1 + _tmp8) - local_point(2, 0) * (-_tmp10 + _tmp9) +
               world_point(1, 0);
  _res(2, 0) = -_world_T_local[6] - local_point(0, 0) * (-_tmp6 + _tmp7) -
               local_point(1, 0) * (_tmp10 + _tmp9) - local_point(2, 0) * (_tmp0 + _tmp8 + 1) +
               world_point(2, 0);

  return _res;
}
\end{lstlisting}
\caption{Generated C++ code for \lstinline{point_residual}}
\label{fig:point-residual-code}
\end{figure}

\subsection{Optimization Framework}
\label{subsec:optimization}

SymForce provides an optimization library in C++ and Python which works naturally with our code generation tools and Lie group types. It performs tangent space optimization using a factor graph formulation inspired by GTSAM \cite{dellaert2012factor} and a low-overhead implementation of the Levenberg-Marquardt algorithm \cite{levenberg1944method, marquardt1963algorithm, transtrum2012improvements}. It is highly optimized for real-time execution, as shown in Sec. \ref{sec:experiments}.

As an example, the \lstinline{point_residual} function can be interpreted as a residual between two points, parameterized by the pose \lstinline{world_T_local}. To minimize this residual, a function is generated that computes the the Jacobian $J$ of the residual $b$, as well as the Gauss-Newton approximation for the Hessian $J^T J$ and right-hand side $J^T b$, which form the Gauss-Newton update $\delta x = -(J^T J) J^T b$. This function is generated automatically from \lstinline{point_residual} using \lstinline{Codegen.with_linearization}. For brevity, we only show the signature:

\begin{lstlisting}[language=C++]
template <typename T>
void PointFactor(
    const sym::Pose3<T>& world_T_local,
    const Eigen::Matrix<T, 3, 1>& world_point,
    const Eigen::Matrix<T, 3, 1>& local_point,
    Eigen::Matrix<T, 3, 1>* res,
    Eigen::Matrix<T, 3, 6>* jacobian,
    Eigen::Matrix<T, 6, 6>* hessian,
    Eigen::Matrix<T, 6, 1>* rhs);
\end{lstlisting}

Note that we could simply output $J$ and $b$ and compute $J^T J$ and $J^T b$ by doing matrix multiplications at runtime, but computing these products symbolically is typically computationally advantageous, as explained in Sec. \ref{subsec:sparsity-exploitation}.

This function is used to construct \lstinline{Factor} objects, which represent residual blocks within the optimization that touch a set of optimized variables. All problem inputs and initial guesses are stored in a \lstinline{Values} class, and the \lstinline{Optimizer} class is invoked to minimize the residual of all factors. The optimizer uses the \lstinline{LieGroupOps} concept on the C++ \lstinline{Values} to perform tangent space retraction.

All of this machinery is also available in Python via a wrapped version of the optimization framework.  This allows for quick prototyping without compiling any code, with the ability to generate C++ from the same symbolic implementation and have confidence that runtime results will be identical.

In the example illustrated above, one instance of the  generated factor is instantiated per measurement in C++. This is very flexible and allows using our library of existing factors or handwritten functions without writing any symbolic code.  However, SymForce also supports generating an entire problem consisting of many residual terms as one large function.  This approach can yield large efficiency gains because of expressions automatically shared between multiple factors, as we show in Sec.  \ref{sec:scan-matching-experiment}.  We provide an \lstinline{OptimizationProblem} class to organize large symbolic problems and generate functions at multiple levels of granularity to feed to the \lstinline{Optimizer}.

Finally, we provide a \lstinline{GncOptimizer} subclass of \lstinline{Optimizer} that implements Graduated Non-Convexity (GNC) \cite{blake1987visual}. GNC is a method for transitioning from a convex cost function to a robust cost function as the optimization converges, to create a wider basin of convergence while still incorporating outlier rejection.  We provide an implementation of the adaptive robust loss function from \cite{barron2019general} using the singularity handling approach described in Sec. \ref{sec:singularity-handling}. Our GNC optimizer works with any tunable loss function.

\section{Speed Advantages}
\label{sec:speed-advantages}

We highlight three ways that symbolic computation speeds up code by reducing the work performed at runtime -- function flattening, sparsity exploitation, and algebraic simplification.

\subsection{Function Flattening}
\label{subsec:function-flattening}

SymForce gains enormous performance advantages by generating runtime functions that consist of a single branchless chain of instructions that share all common subexpressions.

Software engineers strive to organize code into easily composable functions. Often, computing a desired quantity requires invoking many sub-functions. While having structured code improves usability, both the author and the compiler have limited ability to optimize for speed across function boundaries, leading to tension between usability and speed.

Below is a trivial example of a function that uses two helpers, each of which compute common terms inside:

\begin{lstlisting}[language=Python, caption=Function flattening example]
def helper_1(a, b):
    return a**2 + abs(a / b) / b**2

def helper_2(a, b):
    return abs(a / b) + (a**2 - b**2)

def func(a, b):
    return helper_1(a, b) - helper_2(a, b)
\end{lstlisting}

Naively, computing \lstinline{func(a, b)} requires 13 operations and the overhead of two function calls, but a capable compiler could inline these tiny functions and compute the result in just 6 operations, making use of helper variables:

\begin{gather}
    x_{0} = b^{2} \nonumber \\
    x_{1} = \left|{\frac{a}{b}}\right| \\
    x_{0} - x_{1} + \frac{x_{1}}{x_{0}} \nonumber
\end{gather}

In realistic scenarios, most larger functions are too costly for the compiler to inline \cite{241594}, so the execution approaches the naive case. If the redundant calculations and function calls are not acceptable, the alternative is to hand-optimize at the cost of usability by manually flattening the functions or sharing state between the helpers.

Symbolic code addresses this problem with explicit separation between the symbolic and the runtime contexts. The symbolic code is written with small, composable functions, but any evaluated quantities are generated as flat expressions amenable to optimization. In SymForce, computing the runtime variant of \lstinline{func(a, b)} requires 6 operations with no additional work, and the benefits scale to very large expressions.

This process of \textit{flattening} also helps with cache performance, as we demonstrate in detail in Sec. \ref{sec:experiments}.

\subsection{Sparsity Exploitation}
\label{subsec:sparsity-exploitation}

SymForce can yield order of magnitude speedups in the multiplication of matrices that include zero entries. Any amount of sparsity will lead to a large number of terms that do not need to be computed, as they would otherwise be multiplied by zero at runtime.

Take as an example two (6, 6) matrices $X$ and $Y$:

\begin{align}
&X = \begin{bmatrix}
a & 0 & b & 2 b & 0 & 0 \\
0 & a b & 0 & \frac{a}{b} &a^2 & 0 \\
0 & 0 & a b^2 & 0 & \frac{a}{b^2} & 0 \\
\frac{a}{b^3} & 0 & 0 & a b^3 & 0 & \frac{a}{b^4} \\
0 & b^2 & 0 & 0 & a b^4 & 0 \\
0 & 0 & 0 & 0 & 0 & a b^4
\end{bmatrix} \\
&Y = \begin{bmatrix}
0 & -a b & b & 0 & 0 & 0 \\
a b & 0 & -a & 0 & 0 & 0 \\
-b & a & 0 & 0 & 0 & 0 \\
0 & a^2 & 0 & a & 0 & 0 \\
0 & 0 & b^2 & 0 & b & 0 \\
a^2 & 0 & 0 & 0 & 0 & a b
\end{bmatrix}
\end{align}

Dense multiplication consumes $(N + (N-1)) N^2$ scalar operations, or 396 for $N = 6$. Combined with 21 operations to compute the values within the matrices, it takes a total of 417 symbolic operations to compute $X Y$. 

By multiplying matrices symbolically and generating code for the result, we both exploit the structure of the matrices and share expressions between them. $X Y$ can be computed in just 34 symbolic operations, a 12x reduction:

\begin{align}
x_{0} = b^{2}, \;
x_{1} = a b,  \;
x_{2} = a^{2},  \;
x_{3} = b x_{2}, \nonumber \\
x_{4} = x_{0} x_{2},  \;
x_{5} = a^{3},  \;
x_{6} = \frac{1}{b},  \;
x_{7} = b^{3}, \\
x_{8} = a x_{7},  \;
x_{9} = \frac{1}{x_{0}},  \;
x_{10} = b^{6},  \;
x_{11} = b^{5}, \nonumber
\end{align}
\begin{equation}
\scalemath{0.82}{\begin{bmatrix}
-x_0 & x_1 + x_3 & x_1 & 2 x_1 & 0 & 0 \\
x_4 & x_5 x_6 & -x_3 + x_4 & x_2 x_6 & x_3 & 0 \\
-x_8 & x_4 & a & 0 & a x_6 & 0 \\
\frac{x_5}{b^4} & -x_2 x_9 + x_5 x_7 & a x_9 & x_2 x_7 & 0 & \frac{x_2}{x_7} \\
x_8 & 0 & -a x_0 + a x_{10} & 0 & a x_{11} & 0 \\
x_{11} x_5 & 0 & 0 & 0 & 0 & x_{10} x_2
\end{bmatrix}}.
\end{equation}

Beyond the symbolic operation count, memory effects must be considered. Instructions are needed to load inputs into registers, with significant penalties for cache misses. Our method greatly improves cache performance, because the CPU only needs to manage 12 intermediate inputs rather than the 72 entries of the dense matrices. In other words, most entries of $X$ and $Y$ are never represented.


In robotics and computer vision, matrix multiplication is prevalent in transformations, projections, uncertainty propagation, and especially for Jacobians during automatic differentiation. Performance gains compound from longer chains of matrix multiplications and more complex shared terms between them. SymForce can flatten code across thousands of function calls and matrix multiplications into a single branchless function that shares all common subexpressions.

See Sec. \ref{sec:matmul-experiment} for a detailed performance analysis of this key concept across varying matrix sizes and sparsity patterns.

\subsection{Algebraic Simplification}
\label{subsec:algebraic-simplification}

Symbolic expressions can be algebraically simplified into forms that are faster to compute. Categories of simplifications include expansion, factorization, term collection, cancellation, fraction decomposition, trigonometric and logarithmic identities, series expansions, and limits.

SymPy provides a wide array of simplifications. Basic simplifications are done automatically on expression construction, but most require specific invocation, for example with \lstinline{sm.simplify}.\footnote{https://docs.sympy.org/latest/tutorial/simplification.html} While powerful, this technique can require domain expertise and careful effort to achieve notable improvements.




\section{Symbolic Differentiation}
\label{sec:symbolic-differentiation}

In this section we discuss the advantages of symbolic differentiation and present novel techniques for computing tangent-space Jacobians. We demonstrate that users do not have to implement or test any bug-prone handwritten derivatives. In addition, our approach is often faster by avoiding dynamic memory allocation and dense chain ruling at runtime.

We build on tools in SymPy and SymEngine to automatically compute derivatives of vector-space symbolic expressions and extend them to handle our geometry types and \lstinline{Values} class. A key capability provided by SymForce is computing tangent-space derivatives of arbitrary user-defined functions operating on Lie group types, which is necessary for on-manifold optimization and uncertainty propagation.

Sec. \ref{sec:inverse-compose-experiment} shows how these advantages lead to order of magnitude speedups over automatic differentiation.

\subsection{Symbolic vs Automatic Differentiation}
\label{subsec:symbolic-vs-autodiff}
Symbolic differentiation has compelling advantages over automatic differentiation, both by requiring less handwritten code, and by sharing more subexpressions and eliminating the need for matrix multiplication at runtime. 

Automatic differentiation (AD) is the prevalent approach for computing derivatives of large computation graphs \cite{autodiff-review}.  Given a computation graph, AD produces another computation graph to compute a particular derivative, with the size of the resulting graph no bigger than a constant multiple of the original.

It is often claimed that symbolic differentiation is intractable or produces exponentially large computation graphs, and is therefore unusable for nontrivial computations. Consider the chain of function calls  $f(g(h(x, y)))$. The gradient of $f$ with respect to $\begin{bmatrix} x & y \end{bmatrix}$ is expanded as

\begin{equation}
\nabla f = 
\begin{bmatrix}
  \frac{\partial f(g(h(x, y)))}{\partial g(h(x, y))} \frac{\partial g(h(x, y))}{\partial h(x, y)} \frac{\partial h(x, y)}{\partial x} \\
  \frac{\partial f(g(h(x, y)))}{\partial g(h(x, y))} \frac{\partial g(h(x, y))}{\partial h(x, y)} \frac{\partial h(x, y)}{\partial y}
\end{bmatrix}.
\end{equation}

Naively, it appears that $g$ is redundantly evaluated. However, this ignores the use of CSE, which results in one evaluation of each unique function and its derivatives, like in AD.

Furthermore, as described in Sec. \ref{sec:speed-advantages}, representing the derivative as a flattened symbolic expression allows for powerful simplifications across function and matrix multiplication boundaries, for instance in the common case where the Jacobians used by AD contain shared terms or zeros. As a result, our symbolic differentiation and code generation approach outperforms runtime AD for many robotics problems.

\subsection{Tangent-Space Differentiation on Lie Groups}
\label{subsec:tangent-space-diff}
We present two novel methods to automatically and efficiently compute tangent-space derivatives of Lie group elements such as SO(3) and SE(3), leveraging vector-space symbolic differentiation. While we describe the techniques for functions that map Lie groups to $\mathbb{R}^n$, the approach generalizes to functions that output Lie groups.

Lie groups are common parameterizations in robotics and computer vision. When computing a "derivative" of a function whose input is a member of a Lie group, typically the desired quantity is the derivative with respect to a perturbation in the tangent space around the input.  Explicitly, consider a function $f(R)$, $f: SO(3) \rightarrow \mathbb{R}^n$.  Given a retraction operator $R \oplus v$ that applies the perturbation $v \in \mathbb{R}^3$ to $R$, the desired quantity is $\left. \frac{d}{d v}\left[ f(R \oplus v)) \right]\right|_{v=0}$.

In most packages, painstaking care is taken to hand-write these tangent-space derivatives. SymForce computes them all automatically. We provide two approaches for this -- symbolic application of the chain rule and first-order retraction:

\subsubsection{Symbolic Chain Rule Method}
First, it is important to note that while a user of the code operates on Lie group objects, those objects are internally represented as a set of scalar symbols (their ``storage", as described in \ref{subsubsec:geo-types}).  For instance, we represent SO(3) using unit quaternions.  So while the user can implement $f$ using only group operations without knowing about the internals, the expression we build for $f(R)$ is a function of 4 scalars, the quaternion components of $R$.  We define functions $S : SO(3) \rightarrow \mathbb{R}^4$ and $S^{-1} : \mathbb{R}^4 \rightarrow SO(3)$ to map from the manifold object to the storage representation as a vector and back.  If we then let $s = S(R \oplus v)$, we can rewrite the derivative as
\begin{align}
&\tfrac{d}{d v}\left[ f(R \oplus v) \right] \nonumber \\
= &\tfrac{d}{d v}\left[ f(S^{-1}(S(R \oplus v))) \right] \\
= &\tfrac{d}{d s}\left[ f(S^{-1}(s)) \right] \tfrac{d}{d v}\left[ S(R \oplus v))) \right] \nonumber.
\end{align}

\noindent The term $\left. \frac{d}{d v}\left[ S(R \oplus v))) \right] \right|_{v=0}$ on the right is simply the derivative of the storage of $R$ with respect to the perturbation, and does not depend on $f$.  This is a typically simple function of the group elements.  The left term does depend on $f$; but as we noted before, we already have the symbolic representation of $f(S^{-1}(s))$, and we can simply take the symbolic derivative of that expression to get this term.  Notably, neither of these derivatives needs to be handwritten, they can be computed automatically from the form of $f$ and from the other functions specifying the group representation, respectively.  Then the final tangent-space derivative can be computed by symbolically multiplying these two matrices, and generating a flattened expression for runtime.

\subsubsection{First-Order Retraction Method}
Alternatively, we can directly differentiate $f(R \oplus v)$, which is a function between vector spaces, using a first-order approximation of $R \oplus v$ at $v = 0$. This significantly outperforms the previous method in nearly all our trials and is the default approach for computing tangent-space Jacobians in SymForce.

To do this, we first build expressions for the storage entries of $R \oplus v$, which comprise of scalar functions of $v$.  We then substitute each of these expressions for the storage of $R$ in our expression for $f(R)$, producing an expression for $f(R \oplus v)$ which we can then symbolically differentiate.

Because we only care about the behavior at $v = 0$, we can use a first-order approximation of $R \oplus v$ to simplify the expression without loss of correctness.  Explicitly, we use:
\begin{equation}
R \oplus v \approx S^{-1}\left(S(R) + \left. \frac{d}{d v}\left[ S(R \oplus v) \right] \right|_{v=0} v \right),
\end{equation}
which is typically much simpler than $R \oplus v$.

\section{Branching and Singularity Handling}
\label{sec:singularity-handling}

This section describes techniques to avoid branches in algorithmic functions, particularly in the context of handling singularity points. Avoiding branches greatly simplifies routines for manipulating complex expressions, and also has a critical impact on runtime performance.

Symbolic expressions are computation graphs that are separate from the code that builds them. Every symbolic operation has a function, a deterministic number of inputs, and a single output. SymForce does not support arbitrary branching logic within a single expression -- adding conditional statements to Python code will change the structure of the expression being built, but not add conditionals to the generated code.

Instead, many branches can be formulated with primitives like the sign function, absolute value, floor, min, and max. For example, a comparison like \lstinline{(z <= 3) ? a : b} can be represented symbolically as $a + \mathrm{max}(\mathrm{sign}(z - 3), 0) (b - a)$. These operations are performed with bit operations at the assembly level, and do not introduce true branches. As a benefit, this type of branchless programming improves performance because the CPU can pipeline instructions without fear of branch prediction failures \cite{DBLP:journals/corr/abs-1804-00261}.


\subsection{Handling Singularities with Epsilon}
\label{subsec:singularities-epsilon}

We present a novel method for handling removable singularities within symbolic expressions that introduces minimal performance impact by avoiding the need for branching. Functions encountered in robotics are often smooth, but properly addressing singularity points is critical to avoid NaN values.

Consider the function
\begin{equation}
f(x) = \frac{\sin(x)}{x}.
\end{equation}

This function is smooth on its whole domain but has a singularity at $x = 0$, where it takes the form $0/0$.  We can define $f(0) = \lim_{x \rightarrow 0} \frac{\sin(x)}{x} = 1$ and get a smooth function, but the question remains of how to compute this function in a numerically safe way.  A typical approach may be:

\begin{lstlisting}[language=Python, caption=Python example]
def f(x):
    if abs(x) < epsilon:
        # Approximation for small x
        return 1 - x**2 / 6
    else:
        return sin(x) / x
\end{lstlisting}

This has two problems - it does not result in a single symbolic expression for the result, and might introduce a costly branch. Our method is to shift the input to the function away from the singular point with an infinitesimal variable $\epsilon$.  If we assume for a moment that we only care about non-negative $x$, this corresponds to defining a new function
\begin{equation}
\label{eq:f_safe}
f_{\mathrm{safe}}(x) = f(x + \epsilon) = \tfrac{\sin(x + \epsilon)}{x + \epsilon},
\end{equation}
where $\epsilon$ is a small positive constant.

For the general case where $x \in \mathbb{R}$, we first define a function \lstinline{sign_no_zero}, or snz, as 
\begin{equation}
\mathrm{snz}(x) = \begin{cases}
    1  \quad &\text{if} \, x >= 0 \\
    -1 \quad &\text{if} \, x < 0 \\
\end{cases},
\end{equation}

but we can also define it as a branchless expression as

\begin{equation}
\mathrm{snz}(x) = 2\, \mathrm{min}\left(0, \mathrm{sign}(x)\right) + 1.
\end{equation}

Substituting snz into Eq. \ref{eq:f_safe} makes $f_\mathrm{safe}$ valid for $x \in \mathbb{R}$:

\begin{equation}
f_{\mathrm{safe}}(x) = f(x + \mathrm{snz}(x) \epsilon) = \frac{\sin(x + \mathrm{snz}(x) \epsilon)}{x + \mathrm{snz}(x) \epsilon}.
\end{equation}

For a function $f(x)$ with a removable singularity, if $f$ is Lipschitz with constant $M$, it is simple to show that $|| f_{\mathrm{safe}}(x) - f(x) || <= M \epsilon$.  This is typically a perfectly acceptable level of error with a sufficiently small choice of $\epsilon$. The SymForce default epsilon is $2.2\text{e-}15$ for doubles and $1.2\text{e-}6$ for floats, chosen as 10x the machine epsilon.\footnote{The machine epsilon for a floating point type is defined as the smallest number which, when added to 1.0, produces a different number \cite{ieee754}.}

It is worth noting that several common functions do not satisfy the above requirement and have values or derivatives that are not Lipschitz. \lstinline{sqrt} and \lstinline{acos} are defined on $[0, \infty)$ and $[-1, 1]$, respectively, and have infinite derivatives on the boundary. \lstinline{atan2} and \lstinline{abs} have non-removable singularities at 0 in their value and gradient, respectively.  In these cases snz can still perturb the inputs away from the boundary or singularity to prevent unsafe values at runtime, but the general error bound from above does not apply.

In this example the singularity is at $x = 0$, but this approach trivially generalizes to singularities anywhere.

\subsection{Testing for Correctness}
\label{subsec:testing-epsilon}

SymForce can check the correctness of a given symbolic function $f_\mathrm{safe}(x, \epsilon)$ that supposedly uses $\epsilon$ to avoid a removable singularity point at $x = x_0$.

First, the symbolic value of $f_{\mathrm{safe}}(x_0, 0)$ is computed. This should evaluate to an indeterminate form like 0/0, which SymPy will represent as \lstinline{NaN}.  If the value instead is $\pm\infty$, the user is trying to correct for a non-removable singularity. $\epsilon$ cannot correct for this, and an error is returned.

Next, we take $\lim_{x \rightarrow x_0} f_{\mathrm{safe}}(x, 0)$, which is the correct value of the function at the singularity.  We then additionally compute $\lim_{\epsilon \rightarrow 0} f_{\mathrm{safe}}(x_0, \epsilon)$.  This should be a finite value, and it should be equal to the first limit, indicating that the function converges to the correct value for small $\epsilon$.

For many functions, it is crucial that the first derivative is also correct at the singularity. This can be tested automatically with the same strategy.

The typical alternative to our approach, used in e.g. GTSAM and Sophus, is to add branching near the singularity. It is common to locally approximate the point with a Taylor series for this purpose.  A similar approach can be approximated in symbolic code using piecewise functions, but these come at a cost of added complexity and slower performance.

\section{Limitations} 
\label{sec:limitations}
This section describes potential drawbacks to our approach.

The separation between symbolic and runtime contexts requires thinking at a higher level of abstraction than directly writing runtime code. There are many benefits to this approach, but it takes practice. One common error is that conditional statements in Python will not result in branches in runtime code. Users must conceptualize instruction-level branching and employ our concept of epsilon to avoid singularities, as discussed in Sec. \ref{sec:singularity-handling}.

Another common drawback is that our method of flattening expressions becomes impractical with highly nested expressions, such as long chains of integrations or loops. Since the generated code unrolls these chains, it can lead to long compile times, poor cache performance, or other bottlenecks. Further work is required to handle loops or sub-functions as a construct of symbolic expressions. Similar bottlenecks can happen when attempting to generate a linearization function with a very large number of variables. In some cases, it is a better trade-off to generate multiple functions and call them dynamically at runtime, with some sacrifices made in shared subexpressions. SymForce provides tools for exploring these tradeoffs.

Some symbolic routines, like simplification and factorization, become slow with large expressions. We recommend understanding the computational cost of these routines and using them in careful and targeted ways.

SymForce is most suitable for generating functions with up to hundreds of input variables, and hundreds of thousands of instructions. However, functions can be invoked dynamically in much larger optimization problems. SymForce does not directly support data parallelism like operating over pixels of an image, and does not attempt to compete with libraries that do so. However, it can efficiently generate the inner kernel for a single pixel in such a use case.

Finally, SymForce is a young library and there are many things it does not do. For example, our optimizer does not support hard constraints or specialized solvers. However, soft constraints implemented using barrier functions work well and SymForce generated functions can be readily used with other optimization libraries.

\section{Experiments}
\label{sec:experiments}

In this section we present benchmark results on multiple problems implemented with SymForce and alternatives. We compare with Eigen for sparse and dense matrix multiplication, GTSAM, Sophus, and JAX for tangent space differentiation, and GTSAM, Ceres, and JAX for nonlinear least-squares.

We measure CPU time, instruction counts, and L1 cache loads on an Intel i7 CPU and NVIDIA Tegra X2 ARM CPU. All tests are compiled with \lstinline{ -O3 -march=native -ffast-math}, double precision, executing on core 2. Note that while CPU time and L1 cache loads can vary several percent across runs, instruction counts vary by $< 1 \%$ in all experiments.
In addition, we we measure execution time in JAX on an RTX 2080 Ti GPU.
We provide code for all experiments.

\subsection{Matrix Multiplication Experiment}
\label{sec:matmul-experiment}

We first present an experiment demonstrating the performance impact of flattening functions and exploiting sparsity in matrix multiplication, as introduced in Sec. \ref{subsec:function-flattening} and \ref{subsec:sparsity-exploitation}.

We select a series of matrix structures of varying sizes and sparsity from the SuiteSparse Matrix Collection \cite{10.1145/2049662.2049663}. We generate random expressions into the nonzero entries of the matrices from a small set of scalar symbols, using the strategy described by Lample \cite{DBLP:journals/corr/abs-1912-01412}. Our benchmark task is computing $X^T Y$, where $X$ and $Y$ have the same sparsity pattern but with differently generated random expressions. In this example, the expressions are functions of 5 scalar symbols, with approximately 5 operations per expression.


We compare against sparse matrices, dynamic-size dense matrices, and fixed-size dense matrices using the Eigen library in C++. Table \ref{table:matmul2} displays CPU time for computing $X^T Y$ across all matrices and methods.  In each case, there are two function calls to compute $X$ and $Y$ independently, then they are multiplied together. Against these we compare a SymForce flattened function that outputs the product $X^T Y$ directly.

\begin{table*}[t]
\caption{
Benchmark for computing the matrix product $X^T Y$ in C++ for matrices of varying sizes and sparsity patterns.  CPU Time, in nanoseconds, is shown for each method on each matrix. lp\_sc105 failed to fit on the stack for the dense fixed approach.
}
\label{table:matmul2}
\begin{tabularx}{\textwidth}{X|r|r|r|r|r|r} \hline

& b1\_ss & Tina\_DisCog & n3c4\_b2 & bibd\_9\_3 & lp\_sc105 & rotor1 \\
& (7x7), 31\% & (11x11), 40\% & (20x15), 20\% & (36x84), 8\% & (105x163), 2\% & (100x100), 7\% \\
\hline
\multicolumn{6}{c}{Intel Core i7-4790 CPU @ 4GHz, Clang 10.0.1} \\
\hline
Sparse & 1426.9 & 3713.1 & 4264.0 & 30471.3 & 23653.0 & 85965.5 \\
Dense Dynamic & 523.0 & 714.7 & 998.8 & 16834.6 & 345705.8 & 102303.0 \\
Dense Fixed & 108.9 & 547.8 & 815.1 & 16707.6 & N/A & 64874.6 \\
\textbf{SymForce Flattened} & \textbf{22.7} & \textbf{75.0} & \textbf{94.0} & \textbf{9609.0} & \textbf{6165.2} & \textbf{25684.6} \\
SymForce Speedup over Second Best & 4.8x & 7.3x & 8.7x & 1.7x & 3.8x & 2.5x \\
\hline
\multicolumn{6}{c}{Tegra X2 ARM Denver CPU @ 2.035 GHz, GCC 7.5.0} \\
\hline
Sparse & 2738.2 & 6949.8 & 7618.4 & 65411.6 & 51313.9 & 260341.9 \\
Dense Dynamic & 806.9 & 1401.9 & 3443.9 & 97706.6 & 1772193.8 & 559305.4 \\
Dense Fixed & 477.5 & 1178.7 & 2771.7 & 94184.5 & N/A & 337921.8 \\
\textbf{SymForce Flattened} & \textbf{38.1} & \textbf{205.5} & \textbf{239.2} & \textbf{17865.8} & \textbf{9885.4} & \textbf{76670.0} \\
SymForce Speedup over Second Best & 12.5x & 5.7x & 11.6x & 3.7x & 5.2x & 3.4x \\
\end{tabularx}
\end{table*}

We note that the flattened function outperforms the dense fixed approach in all categories. We leverage the benefits of sparse multiplication, but without the memory and index management overhead it introduces.

The results also show that for small matrices, dynamic memory allocation is a poor tradeoff and dominates the computational time. Dense multiplication foregoes any benefits of sparsity, but better leverages SIMD instructions.

Table \ref{table:matmul} shows detailed results for the matrix  n3c4\_b2, which is 20 x 15 and has 20\% sparsity. Our method outperforms the second best by 8.7x on Intel and 11.6x on Tegra. One dramatic difference is in the number of L1 data cache loads required. For this small size, most of the variables are simply kept on CPU registers. The CPU only needs to manage intermediate inputs rather than all the entries of the dense matrices. In other words, most entries of $X$ and $Y$ are never explicitly held in registers.

\begin{table}
\caption{
Detailed results for the matrix n3c4\_b2, which is 20 x 15 and has 20\% sparsity. Measured with the linux \lstinline{perf} tool, amortized across one million runs.  Here we also show the time required by JAX, both on the CPU and GPU.  Even amortized over a batch size of 1000, a large amount of overhead remains.
}
\label{table:matmul}
\begin{tabularx}{\columnwidth}{X|r|r|r} \hline
                  & Time (ns) & Instructions & L1 Loads \\ \hline
\multicolumn{4}{c}{Intel Core i7-4790 CPU @ 4GHz, Clang 10.0.1} \\ \hline
      JAX, Batch = 1000 &  3205.4 & N/A & N/A \\
            Sparse & 4264.0 &  42755 &  14185 \\
     Dense Dynamic &  998.8 &   9812 &   2777 \\
      Dense Fixed &  815.1 &    8182 &   2671 \\
\textbf{SymForce Flattened} & \textbf{94.0}  & \textbf{870} & \textbf{225} \\ \hline
\multicolumn{4}{c}{Tegra X2 ARM Denver CPU @ 2.035 GHz, GCC 7.5.0} \\ \hline
            Sparse & 7618.4 &  29030 &  98439 \\
     Dense Dynamic &  3443.9 &   13737 & 28627 \\
      Dense Fixed &  2771.7 &    10735 &  2243 \\
\textbf{SymForce Flattened} & \textbf{239.2}  & \textbf{892} & \textbf{278} \\ \hline
\multicolumn{4}{c}{RTX 2080 Ti GPU @ 1850MHz} \\ \hline
      JAX, Batch = 1000 &  6699.3 & N/A & N/A \\ \hline
\end{tabularx}
\end{table}

\subsection{Inverse Compose Experiment}
\label{sec:inverse-compose-experiment}

We now extend the idea of the previous experiment into a practical example of computing tangent space Jacobians. Consider transforming a point by the inverse of a pose, a common operation in SLAM or bundle adjustment problems. This expression can be computed by invoking two functions - inversion and composition:

\smallskip

\lstinline{result = pose.inverse() * point}

\smallskip

The automatic differentiation approach to compute the Jacobian of this expression with respect to the tangent space of the pose is to chain the Jacobians of the inverse and compose operations. In pseudocode, the operations that actually happen at runtime  might look like:

\begin{lstlisting}[language=C++]
inverse, inverse_D_pose = InverseWithJacobian(pose);
res, res_D_inverse = ComposeWithJacobian(
    inverse, point);
res_D_pose = res_D_inverse * inverse_D_pose;
\end{lstlisting}

This approach is typical in robotics because it leverages existing Jacobian implementations for the underlying functions, and avoids handwriting a Jacobian for the combined function. However, as described in Sec. \ref{sec:speed-advantages}, this sacrifices performance because it misses the opportunity to share common subexpressions between the operations and requires dense multiplication of the matrices at runtime.

We use SymForce to generate Jacobian functions for the inverse and compose functions independently, and compare to a flattened function that fuses them together. We can estimate the runtime from the number of symbolic operations required, using the SymPy \lstinline{count_ops} function after performing CSE: 

\smallskip\smallskip
\noindent \begin{tabularx}{\columnwidth}{X|r} \hline
                                    Function &   Operations \\ \hline
                          Inverse + Jacobian &           73 \\
                          Compose + Jacobian &           85 \\
  $(3, 6)$ by $(6, 6)$ Matrix Multiplication &          198 \\ \hline
                       Total for AD Approach &          356 \\ \hline
        Flattened Inverse Compose + Jacobian &   \textbf{91} \\ \hline
\end{tabularx}
\smallskip

\noindent By this heuristic, the flattened expression (91 ops) should be approximately 4x faster than the AD approach (356 ops). This ignores memory effects, but our approach also reduces the number of intermediate terms accessed.

\begin{table}
    \caption{Benchmark results for the inverse compose experiment. * - GTSAM includes a handwritten derivative for this operation.}
    \label{table:inverse_pose_benchmark}
    \begin{tabularx}{\columnwidth}{X|r|r|r} \hline
                   & Time (ns) & Instructions & L1 Loads \\ \hline
\multicolumn{4}{c}{Intel Core i7-9700K CPU @ 4.6GHz, Clang 10.0.0} \\ \hline
       JAX, Batch = 1000 &  1950.0 & N/A & N/A \\
    Sophus Chained & 139.4 &   1904.4 &  443.6 \\
     GTSAM Chained &  74.9 &    922.2 &  295.6 \\
  SymForce Chained &  42.9 &   448.1 &   127.6 \\
      GTSAM Custom* &  11.3 &    \textbf{102.1} &   36.6 \\
\textbf{SymForce Flattened} & \textbf{7.9}  & 134.0 & \textbf{26.6} \\ \hline
\multicolumn{4}{c}{Tegra X2 ARM Denver CPU @ 2.035 GHz, GCC 7.5.0} \\ \hline
    Sophus Chained &  360.6 &   482.4 &  294.9 \\
     GTSAM Chained &  230.8 &  690.1 &  312.4 \\
  SymForce Chained &  106.6 &   372.8 &  107.3 \\
      GTSAM Custom* &   58.7 &    133.8 &  77.6 \\
\textbf{SymForce Flattened} & \textbf{20.3}  & \textbf{105.7} & \textbf{23.0} \\ \hline
\multicolumn{4}{c}{RTX 2080 Ti @ 1850MHz} \\ \hline
       JAX, Batch = 1000 &  6642.4 & N/A & N/A \\ \hline
\end{tabularx}
\end{table}

We present benchmark results comparing automatic differentiation (chaining) with our flattened expressions in Table~\ref{table:inverse_pose_benchmark}.

First, we note that the chained versions are significantly slower than the flattened versions. We also note that the SymForce chained function is faster than alternative chained versions, due to the efficient implementations of the autogenerated geometry types.

``GTSAM Custom" refers to the function \lstinline{gtsam::Pose3::transform_to}, which provides a handwritten value and Jacobian for the inverse compose operation, separately from the inverse or compose functions used in the chaining method. The SymForce generated flattened function outperforms the handwritten method, with no specialization required. In the case of the Intel CPU, the GTSAM Custom function has fewer instructions but the SymForce Flattened function has higher speed and better cache performance.

Note that for larger expressions, there will typically not preexist a handwritten method with Jacobians.  It is often difficult or impossible to implement handwritten Jacobians for large functions, and doing so consumes valuable engineering time.  In the general case, it is fair to compare the autogenerated SymForce Flat version with the chained approaches of alternative libraries, which in this case yields 9.5x and 11.5x improvements on the Intel and Tegra CPUs.

\subsection{Robot 3D Localization Example}
\label{sec:scan-matching-experiment}

\begin{figure}
    \centering
    \includegraphics[width=\columnwidth]{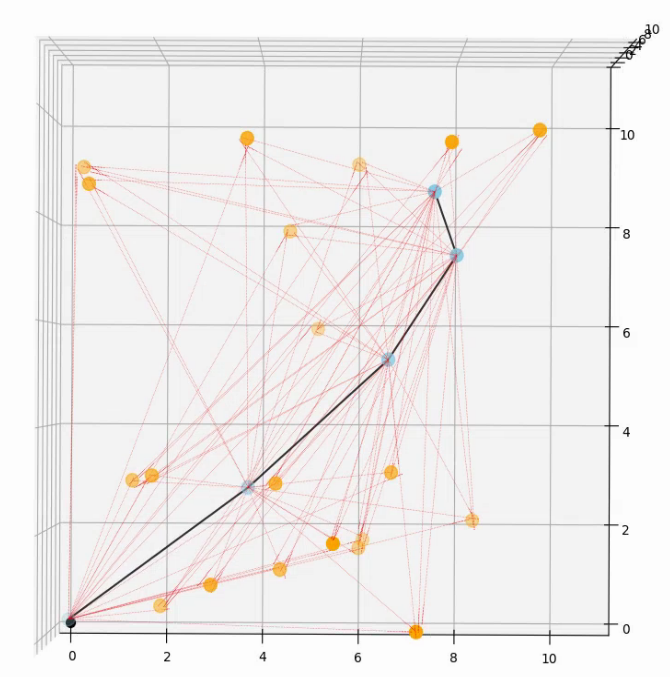}
    \caption{Robot Localization example.  This shows the converged solution, with robot poses in blue, and landmarks in orange.}
    \label{fig:robot-localization}
    \vspace{2em}
\end{figure}

Next, we show a practical example of estimating a robot trajectory given point measurements.  As shown in Figure~\ref{fig:robot-localization}, we set up a sequence of 5 poses.  The environment contains 20 point landmarks at known world positions.  At each timestep, the robot measures the position of each landmark in its local frame and measures its odometry from the previous timestep as a relative pose, both with Gaussian noise.  We initialize the poses at the origin and optimize the poses to convergence.

\begin{table}
    \vspace{1em}
    \caption{Results for the Robot Localization example. The experimental setup is exactly as detailed in the caption of Table \ref{table:matmul}. Linearize is the time to evaluate the residual, Jacobian, and Hessian of the problem. Iterate includes linearization, solving, and updating the variables. All optimizers show similar convergence at around 12 iterations, so we focus on the timing for a single iteration.}
    \label{table:robot-localization}
    \begin{tabularx}{\columnwidth}{X|r|r|r} \hline
                   & Time (us) & Instructions & L1 Loads \\ \hline
\multicolumn{4}{c}{Intel Core i7-9700K CPU @ 4.6GHz, Clang 10.0.0} \\ \hline
            Linearize: Ceres  & 42.6 &  526.0 &  151.0 \\
            Linearize: JAX, Batch = 1000  & 35.4 & N/A & N/A \\
            Linearize: GTSAM &  30.0 &   351.7 &   87.6 \\
  Linearize: SymForce Dynamic &  15.1 &   220.4 &   54.3 \\
\textbf{Linearize: SymForce Fixed} & \textbf{5.4}  & \textbf{43.2} & \textbf{17.4} \\ \hline
\multicolumn{4}{c}{Tegra X2 ARM Denver CPU @ 2.035 GHz, GCC 7.5.0} \\ \hline
            Linearize: Ceres  & 591.1 &  1203.9 &  517.2 \\
            Linearize: GTSAM &  190.0 &   572.0 &   186.7 \\
 Linearize: SymForce Dynamic  &  73.6 &   192.1 &   61.6 \\
\textbf{Linearize: SymForce Fixed } & \textbf{40.8}  & \textbf{60.4} & \textbf{23.3} \\ \hline
\multicolumn{4}{c}{RTX 2080 Ti @ 1850MHz} \\ \hline
       Linearize: JAX, Batch = 1000 &  8.9 & N/A & N/A \\ \hline
\multicolumn{4}{c}{Intel Core i7-9700K CPU @ 4.6GHz, Clang 10.0.0} \\ \hline
            Iterate: Ceres  & 108.6 &  1152.5 &  315.0 \\
            Iterate: GTSAM  & 155.4 &   1713.4 &   427.8 \\
 Iterate: SymForce Dynamic  &  47.0 &    587.2 &   151.6 \\
\textbf{Iterate: SymForce Fixed } & \textbf{25.6}  & \textbf{223.0} & \textbf{73.5} \\ \hline
\multicolumn{4}{c}{Tegra X2 ARM Denver CPU @ 2.035 GHz, GCC 7.5.0} \\ \hline
            Iterate: Ceres  & 967.8 &  1809.5 &  747.6 \\
            Iterate: GTSAM &  557.6 &  1289.7 &  438.5 \\
 Iterate: SymForce Dynamic  & 231.6 &   551.5 &   189.1 \\
\textbf{Iterate: SymForce Fixed } & \textbf{166.1}  & \textbf{272.9} & \textbf{107.3} \\ \hline
    \end{tabularx}
\end{table}

We show timing results in Table~\ref{table:robot-localization}.  The ``Dynamic" formulation generates one factor per measurement and builds a factor graph at runtime.  This allows for dynamically sized problems, where the number of poses is not known at code generation time or may change over time.  The ``Fixed" formulation represents generating a single function to linearize the entire problem.  This requires that the size of the problem is fixed at code generation time.  However, we demonstrate significant performance gains in cases where this is possible, due to common subexpression elimination between factors.

We note that SymForce is the fastest of the dynamic variants, due to our efficient linearization function and geometry types. In addition, our fixed variant significantly outperforms our dynamic variant. This demonstrates that for smaller problems, it is very compelling to flatten the entire linearization into a single function that shares all computation.

We also note that the GPU time for JAX is competitive at a batch size of 1000. In contrast to previous problems, this example is large enough to amortize the overhead. However, in many cases it is impractical to run with a batch size of 1000, and the per-instance evaluation slows down massively with smaller batch sizes.






\section{Conclusion} 
\label{sec:conclusion}


In this paper we presented SymForce and its underlying approaches as a powerful strategy for solving a range of computational problems. For its primary domain of real-time robotics applications, SymForce results in faster runtime code, sometimes by an order of magnitude, while requiring less development time and fewer lines of handwritten code versus state-of-the-art alternatives. Its geometry types, symbolic manipulation tools, code generator, and tangent-space optimization machinery work together to provide a workflow from prototyping of ideas to running in production. Its key performance advantages come from autogenerating code that flattens across function calls and matrix multiplications, taking advantage of all common subexpressions and any amount of problem sparsity, particularly avoiding unnecessary operations that come from the chain rule in automatic differentiation.



\section*{Acknowledgments}
\label{sec:acknowledgements}

The authors are grateful to Skydio for supporting our work, and to all who provided feedback on it. Thank you to the SymPy and SymEngine community for building a strong foundation, and to Frank Dellaert for his work on GTSAM, which motivated several design elements of our work.


\bibliographystyle{unsrtnat}
\bibliography{references}

\begin{thebibliography}{35}
\providecommand{\natexlab}[1]{#1}
\providecommand{\url}[1]{\texttt{#1}}
\expandafter\ifx\csname urlstyle\endcsname\relax
  \providecommand{\doi}[1]{doi: #1}\else
  \providecommand{\doi}{doi: \begingroup \urlstyle{rm}\Url}\fi

\bibitem[Meurer et~al.(2017)Meurer, Smith, Paprocki, \v{C}ert\'{i}k, Kirpichev,
  Rocklin, Kumar, Ivanov, Moore, Singh, Rathnayake, Vig, Granger, Muller,
  Bonazzi, Gupta, Vats, Johansson, Pedregosa, Curry, Terrel, Rou\v{c}ka, Saboo,
  Fernando, Kulal, Cimrman, and Scopatz]{10.7717/peerj-cs.103}
Aaron Meurer, Christopher~P. Smith, Mateusz Paprocki, Ond\v{r}ej
  \v{C}ert\'{i}k, Sergey~B. Kirpichev, Matthew Rocklin, Amit Kumar, Sergiu
  Ivanov, Jason~K. Moore, Sartaj Singh, Thilina Rathnayake, Sean Vig, Brian~E.
  Granger, Richard~P. Muller, Francesco Bonazzi, Harsh Gupta, Shivam Vats,
  Fredrik Johansson, Fabian Pedregosa, Matthew~J. Curry, Andy~R. Terrel,
  \v{S}t\v{e}p\'{a}n Rou\v{c}ka, Ashutosh Saboo, Isuru Fernando, Sumith Kulal,
  Robert Cimrman, and Anthony Scopatz.
\newblock {SymPy}: symbolic computing in python.
\newblock \emph{PeerJ Computer Science}, 3:\penalty0 e103, January 2017.
\newblock ISSN 2376-5992.
\newblock URL \url{https://sympy.org}.

\bibitem[Guennebaud et~al.(2010)Guennebaud, Jacob, et~al.]{eigenweb}
Ga\"{e}l Guennebaud, Beno\^{i}t Jacob, et~al.
\newblock Eigen v3, 2010.
\newblock URL \url{http://eigen.tuxfamily.org}.

\bibitem[Margossian(2019)]{autodiff-review}
Charles~C. Margossian.
\newblock A review of automatic differentiation and its efficient
  implementation.
\newblock \emph{WIREs Data Mining and Knowledge Discovery}, 9\penalty0 (4), Mar
  2019.
\newblock ISSN 1942-4795.

\bibitem[Maplesoft(2022)]{maple}
Maplesoft.
\newblock Maple, 2022.
\newblock URL \url{https://www.maplesoft.com/products/maple/}.

\bibitem[Inc.(2021)]{Mathematica}
Wolfram~Research{,} Inc.
\newblock Mathematica, {V}ersion 13.0.0, 2021.
\newblock URL \url{https://www.wolfram.com/mathematica}.

\bibitem[Inc.(2022)]{matlabsymbolic}
The~MathWorks Inc.
\newblock Matlab symbolic math toolbox, 2022.
\newblock URL \url{https://www.mathworks.com/products/symbolic.html}.

\bibitem[Aln\ae{}s and Mardal(2010)]{10.1145/1644001.1644007}
Martin~Sandve Aln\ae{}s and Kent-Andr\'{e} Mardal.
\newblock On the efficiency of symbolic computations combined with code
  generation for finite element methods.
\newblock 37\penalty0 (1), jan 2010.
\newblock ISSN 0098-3500.
\newblock \doi{10.1145/1644001.1644007}.

\bibitem[Gede et~al.(2013)Gede, Peterson, Nanjangud, Moore, and
  Hubbard]{10.1115/DETC2013-13470}
Gilbert Gede, Dale~L. Peterson, Angadh~S. Nanjangud, Jason~K. Moore, and Mont
  Hubbard.
\newblock Constrained multibody dynamics with {Python}: From symbolic equation
  generation to publication.
\newblock In \emph{International Conference on Multibody Systems, Nonlinear
  Dynamics, and Control}, 2013.
\newblock URL \url{https://www.pydy.org/}.

\bibitem[Hereid and Ames(2017)]{frost}
Ayonga Hereid and Aaron~D. Ames.
\newblock {FROST}: Fast robot optimization and simulation toolkit.
\newblock In \emph{International Conference on Intelligent Robots and Systems
  (IROS)}, pages 719--726, 2017.

\bibitem[Čertík et~al.(2022)Čertík, Fernando, et~al.]{symengine}
Ondřej Čertík, Isuru Fernando, et~al.
\newblock {SymEngine}, 2022.
\newblock URL \url{https://github.com/symengine/symengine}.

\bibitem[Fernando(2016)]{symengine-speedups}
Isuru Fernando.
\newblock {SymEngine} and code generation, 2016.
\newblock URL
  \url{https://github.com/isuruf/symengine/wiki/SymEngine-and-code-generation}.

\bibitem[Dellaert(2012)]{dellaert2012factor}
Frank Dellaert.
\newblock Factor graphs and {GTSAM}: A hands-on introduction.
\newblock Technical report, Georgia Institute of Technology, 2012.
\newblock URL \url{https://gtsam.org/}.

\bibitem[Agarwal et~al.()Agarwal, Mierle, et~al.]{ceres-solver}
Sameer Agarwal, Keir Mierle, et~al.
\newblock Ceres solver.
\newblock URL \url{http://ceres-solver.org}.

\bibitem[Kümmerle et~al.(2011)Kümmerle, Grisetti, Strasdat, Konolige, and
  Burgard]{5979949}
Rainer Kümmerle, Giorgio Grisetti, Hauke Strasdat, Kurt Konolige, and Wolfram
  Burgard.
\newblock g2o: A general framework for graph optimization.
\newblock In \emph{International Conference on Robotics and Automation}, pages
  3607--3613, 2011.

\bibitem[Strasdat(2021)]{sophus}
Hauke Strasdat.
\newblock Sophus, 2021.
\newblock URL \url{https://github.com/strasdat/Sophus}.

\bibitem[Deray and Solà(2020)]{Deray-20-JOSS}
Jérémie Deray and Joan Solà.
\newblock Manif: A micro {L}ie theory library for state estimation in robotics
  applications.
\newblock \emph{Journal of Open Source Software}, 5\penalty0 (46):\penalty0
  1371, 2020.
\newblock URL \url{https://github.com/artivis/manif}.

\bibitem[Koppel and Waslander(2018)]{koppel2018manifold}
Leonid Koppel and Steven~L. Waslander.
\newblock Manifold geometry with fast automatic derivatives and coordinate
  frame semantics checking in {C++}.
\newblock In \emph{15th Conference on Computer and Robot Vision (CRV)}, 2018.

\bibitem[Teed and Deng(2021)]{DBLP:journals/corr/abs-2103-12032}
Zachary Teed and Jia Deng.
\newblock Tangent space backpropagation for 3d transformation groups.
\newblock \emph{CoRR}, abs/2103.12032, 2021.

\bibitem[Paszke et~al.(2019)Paszke, Gross, Massa, Lerer, Bradbury, Chanan,
  Killeen, Lin, Gimelshein, Antiga, Desmaison, Kopf, Yang, DeVito, Raison,
  Tejani, Chilamkurthy, Steiner, Fang, Bai, and Chintala]{NEURIPS2019_9015}
Adam Paszke, Sam Gross, Francisco Massa, Adam Lerer, James Bradbury, Gregory
  Chanan, Trevor Killeen, Zeming Lin, Natalia Gimelshein, Luca Antiga, Alban
  Desmaison, Andreas Kopf, Edward Yang, Zachary DeVito, Martin Raison, Alykhan
  Tejani, Sasank Chilamkurthy, Benoit Steiner, Lu~Fang, Junjie Bai, and Soumith
  Chintala.
\newblock {PyTorch}: An imperative style, high-performance deep learning
  library.
\newblock In \emph{Advances in Neural Information Processing Systems 32}, pages
  8024--8035. 2019.
\newblock URL \url{https://pytorch.org/}.

\bibitem[Abadi et~al.(2015)Abadi, Agarwal, Barham, Brevdo, Chen, Citro,
  Corrado, Davis, Dean, Devin, Ghemawat, Goodfellow, Harp, Irving, Isard, Jia,
  Jozefowicz, Kaiser, Kudlur, Levenberg, Man\'{e}, Monga, Moore, Murray, Olah,
  Schuster, Shlens, Steiner, Sutskever, Talwar, Tucker, Vanhoucke, Vasudevan,
  Vi\'{e}gas, Vinyals, Warden, Wattenberg, Wicke, Yu, and
  Zheng]{tensorflow2015-whitepaper}
Mart\'{\i}n Abadi, Ashish Agarwal, Paul Barham, Eugene Brevdo, Zhifeng Chen,
  Craig Citro, Greg~S. Corrado, Andy Davis, Jeffrey Dean, Matthieu Devin,
  Sanjay Ghemawat, Ian Goodfellow, Andrew Harp, Geoffrey Irving, Michael Isard,
  Yangqing Jia, Rafal Jozefowicz, Lukasz Kaiser, Manjunath Kudlur, Josh
  Levenberg, Dandelion Man\'{e}, Rajat Monga, Sherry Moore, Derek Murray, Chris
  Olah, Mike Schuster, Jonathon Shlens, Benoit Steiner, Ilya Sutskever, Kunal
  Talwar, Paul Tucker, Vincent Vanhoucke, Vijay Vasudevan, Fernanda Vi\'{e}gas,
  Oriol Vinyals, Pete Warden, Martin Wattenberg, Martin Wicke, Yuan Yu, and
  Xiaoqiang Zheng.
\newblock {TensorFlow}: Large-scale machine learning on heterogeneous systems,
  2015.
\newblock URL \url{https://www.tensorflow.org/}.

\bibitem[Bradbury et~al.(2018)Bradbury, Frostig, Hawkins, Johnson, Leary,
  Maclaurin, Necula, Paszke, Vander{P}las, Wanderman-{M}ilne, and
  Zhang]{jax2018github}
James Bradbury, Roy Frostig, Peter Hawkins, Matthew~James Johnson, Chris Leary,
  Dougal Maclaurin, George Necula, Adam Paszke, Jake Vander{P}las, Skye
  Wanderman-{M}ilne, and Qiao Zhang.
\newblock {JAX}: composable transformations of {P}ython+{N}um{P}y programs.
\newblock 2018.
\newblock URL \url{http://github.com/google/jax}.

\bibitem[White and Lubin(2021)]{juliadiff}
Lyndon White and Miles Lubin.
\newblock Juliadiff, 2021.
\newblock URL \url{https://juliadiff.org/}.

\bibitem[Lam et~al.(2015)Lam, Pitrou, and Seibert]{numba}
Siu~Kwan Lam, Antoine Pitrou, and Stanley Seibert.
\newblock Numba: a {LLVM}-based python {JIT} compiler.
\newblock In \emph{Proceedings of the Second Workshop on the LLVM Compiler
  Infrastructure in HPC}, LLVM '15, New York, NY, USA, 2015. Association for
  Computing Machinery.
\newblock ISBN 9781450340052.

\bibitem[Gregor et~al.(2006)Gregor, J\"{a}rvi, Siek, Stroustrup, Dos~Reis, and
  Lumsdaine]{10.1145/1167515.1167499}
Douglas Gregor, Jaakko J\"{a}rvi, Jeremy Siek, Bjarne Stroustrup, Gabriel
  Dos~Reis, and Andrew Lumsdaine.
\newblock Concepts: Linguistic support for generic programming in c++.
\newblock \emph{SIGPLAN Not.}, 41\penalty0 (10):\penalty0 291–310, oct 2006.
\newblock ISSN 0362-1340.

\bibitem[Ronacher et~al.(2022)Ronacher, Lord, et~al.]{jinja}
Armin Ronacher, David Lord, et~al.
\newblock Jinja, 2022.
\newblock URL \url{https://github.com/pallets/jinja}.

\bibitem[Levenberg(1944)]{levenberg1944method}
Kenneth Levenberg.
\newblock A method for the solution of certain non-linear problems in least
  squares.
\newblock \emph{Quarterly of applied mathematics}, 2\penalty0 (2):\penalty0
  164--168, 1944.

\bibitem[Marquardt(1963)]{marquardt1963algorithm}
Donald~W Marquardt.
\newblock An algorithm for least-squares estimation of nonlinear parameters.
\newblock \emph{Journal of the society for Industrial and Applied Mathematics},
  11\penalty0 (2):\penalty0 431--441, 1963.

\bibitem[Transtrum and Sethna(2012)]{transtrum2012improvements}
Mark~K Transtrum and James~P Sethna.
\newblock Improvements to the levenberg-marquardt algorithm for nonlinear
  least-squares minimization.
\newblock \emph{arXiv preprint arXiv:1201.5885}, 2012.

\bibitem[Blake and Zisserman(1987)]{blake1987visual}
Andrew Blake and Andrew Zisserman.
\newblock \emph{Visual reconstruction}.
\newblock MIT press, 1987.

\bibitem[Barron(2019)]{barron2019general}
Jonathan~T Barron.
\newblock A general and adaptive robust loss function.
\newblock In \emph{Conference on Computer Vision and Pattern Recognition},
  pages 4331--4339, 2019.

\bibitem[Chen et~al.(1993)Chen, Chang, Conte, and Hwu]{241594}
W.Y. Chen, P.P. Chang, T.M. Conte, and W.W. Hwu.
\newblock The effect of code expanding optimizations on instruction cache
  design.
\newblock \emph{IEEE Transactions on Computers}, 42\penalty0 (9):\penalty0
  1045--1057, 1993.
\newblock \doi{10.1109/12.241594}.

\bibitem[Mittal(2018)]{DBLP:journals/corr/abs-1804-00261}
Sparsh Mittal.
\newblock A survey of techniques for dynamic branch prediction.
\newblock \emph{CoRR}, abs/1804.00261, 2018.

\bibitem[iee(2019)]{ieee754}
{IEEE} standard for floating-point arithmetic.
\newblock \emph{IEEE Std 754-2019 (Revision of IEEE 754-2008)}, pages 1--84,
  2019.
\newblock \doi{10.1109/IEEESTD.2019.8766229}.

\bibitem[Davis and Hu(2011)]{10.1145/2049662.2049663}
Timothy~A. Davis and Yifan Hu.
\newblock The university of florida sparse matrix collection.
\newblock \emph{ACM Trans. Math. Softw.}, 38\penalty0 (1), dec 2011.
\newblock ISSN 0098-3500.
\newblock URL \url{https://sparse.tamu.edu/}.

\bibitem[Lample and Charton(2019)]{DBLP:journals/corr/abs-1912-01412}
Guillaume Lample and Fran{\c{c}}ois Charton.
\newblock Deep learning for symbolic mathematics.
\newblock \emph{CoRR}, abs/1912.01412, 2019.

\end{thebibliography}




\end{document}